\pdfoutput=1
\documentclass[11pt]{article}

\usepackage{acl}

\usepackage{times}
\usepackage{latexsym}

\usepackage[T1]{fontenc}

\usepackage[utf8]{inputenc}

\usepackage{microtype}

\title{Aspect-based Analysis of Advertising Appeals \\ for Search Engine Advertising}

\usepackage{multirow}
\usepackage{booktabs}
\usepackage{arydshln}
\usepackage{pdfpages}
\usepackage{graphicx}
\usepackage{url}
\usepackage{bm}
\usepackage{amsfonts}

\newcounter{num}

\usepackage{CJKutf8}
\usepackage[utf8]{inputenc}
\usepackage[T1]{fontenc}
\newcommand{\ja}[1]{\begin{CJK}{UTF8}{ipxm}#1\end{CJK}}

\usepackage{color}
\newcommand{\draft}[1]{\textcolor{black}{#1}}
\newcommand{\ready}[1]{\textcolor{black}{#1}}

\usepackage[subrefformat=parens]{subcaption}

\author{
    Soichiro Murakami$^{1,2}$, \ Peinan Zhang$^1$, \ Sho Hoshino$^1$, Hidetaka Kamigaito$^2$, \\
    {\bf Hiroya Takamura}$^2$, {\bf Manabu Okumura}$^2$ \\
  $^1$CyberAgent, Inc., $^2$Tokyo Institute of Technology \\
  {\tt 	\{murakami\_soichiro,hoshino\_sho,zhang\_peinan\}@cyberagent.co.jp}, \\
  {\tt kamigaito@lr.pi.titech.ac.jp}, {\tt \{takamura,oku\}@pi.titech.ac.jp} \\
}

\begin{document}
\maketitle
\begin{abstract}
Writing an ad text that attracts people and persuades them to click or act is essential for the success of search engine advertising. Therefore, ad creators must consider various aspects of advertising appeals (A$^3$) such as the {\it price}, {\it product features}, and {\it quality}. 
However, products and services exhibit unique effective A$^3$ for different industries.
In this work, we focus on exploring the effective A$^3$ for different industries with the aim of assisting the ad creation process. 
To this end, we created a dataset of advertising appeals and used an existing model that detects various aspects for ad texts.
Our experiments demonstrated 
that different industries have their own effective A$^3$ and that the identification of the A$^3$ contributes to the estimation of advertising performance.
\end{abstract}

\section{Introduction}

\begin{figure}[t]
\centering
\includegraphics[width=1.0\linewidth]{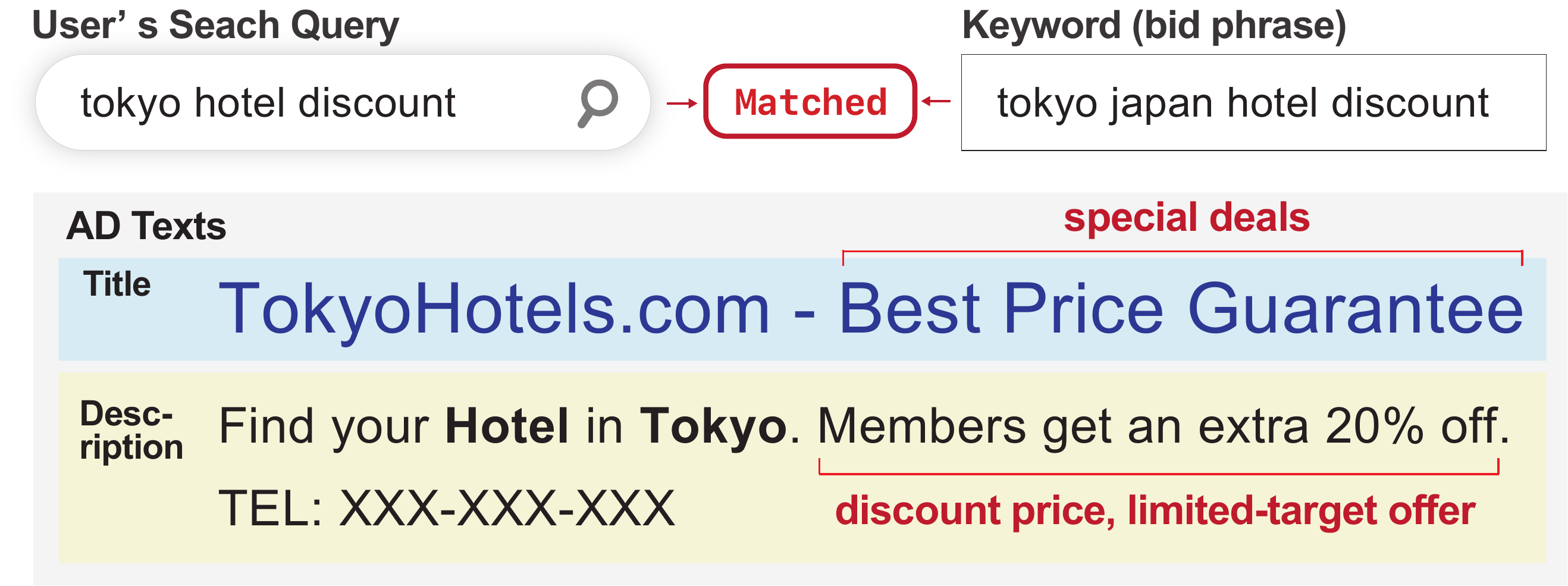}
\caption{Example ad text and its corresponding A$^3$.\label{fig:example_of_td}}
\end{figure}

Search engine advertising (SEA) displays an ad text that consists of a title and a description that are relevant to search queries in search engines, as illustrated in Figure \ref{fig:example_of_td}.
SEA plays an important role in sales promotion and marketing as it allows advertisers to approach users who are interested in specific search queries effectively \cite{fain2006a-brief-history}.
Ad creators write an ad text that attracts the attention of users and persuades them to click or act by introducing various aspects of advertising appeals (denoted as A$^3$ in this paper for short), such as {\it special deals}, as shown in Figure \ref{fig:example_of_td}.
However, products and services exhibit unique effective A$^3$ for different industries.
For example, {\it limited offers} may be attractive to users in the e-commerce (EC) industry, whereas the {\it quality of products} may be more important in the automobile industry.

Thus, we argue that the suggestion of effective A$^3$ for various industries can offer assistance to ad creators.
Therefore, we need to discover the effective aspects.
However, although aspect-based text analysis has attracted significant attention in the review analysis for products and services \cite{akhtar2017-dk,chen2019-tl}, it has received less focus in the advertisement field.

In this work, to deal with this problem, we defined the A$^3$ and constructed a dataset of ad texts that are annotated with A$^3$ in various industries as a first attempt towards assisting ad creators with A$^3$.
Subsequently we developed an aspect detection model to identify different A$^3$ and performed correlation analysis between A$^3$ and the click-through rate (CTR), which is used for supporting ad creation, as an advertising performance metric to explore the effective aspects in different industries.
Furthermore, we investigated the effectiveness of A$^3$ in CTR prediction as a potential application for ad creation support.

Through correlation analysis in our experiments, we found that different industries exhibit unique effective A$^3$.
Furthermore, we found that the identification of the A$^3$ contributes to the CTR prediction.

\section{Related Work}
\paragraph{Ad Creation Support}
Attempts have been made to perform automatic generation of ad texts and keywords \cite{ravi2010-cc,hughes2019kdd,kamigaito-etal-2021-empirical} as well as the estimation of advertising performance metrics such as the CTR \cite{richardson2007-nx,zhang2014-kd,mishra2021tsi} to support the ad creation process.
In this work, we tackle the discovery of the effective A$^3$ for various industries and apply the A$^3$ to CTR prediction with the goal of improving the efficiency of the ad creation process.

\paragraph{Aspect-based Text Analysis}
Although aspect-based text analysis has attracted significant attention, the majority of studies have been limited to specific domains such as hotels, restaurants, and home appliances  \cite{pontiki2016-yz,akhtar2017-dk,chen2019-tl}.
Moreover, as the product review analysis focuses on the aspects of each product, the defined aspects are extremely fine grained (e.g., the {\it modes}, {\it energy efficiency}, and {\it noise} for refrigerators \cite{li2020-up}). 
These aspects are not suitable for ad creation because ad creators must deal with ad texts for various products in multiple industries. 
Therefore, ad creators are required to consider numerous A$^3$. 
In this study, we carefully designed labels that cover the A$^3$ for the general purpose of exploring these in a wide range of industries.
Furthermore, we explored methods for aspect detection, as in the previous work \cite{bagheri2013-jl}, as well as the identification of the effective aspects in terms of advertising performance metrics such as the CTR. 

\section{Construction of A$^3$ Dataset\label{sec:annotation_work}}
\subsection{Data Collection}
\begin{table}[]
\centering
{\small
\begin{tabular}{@{}l@{ }l@{\hspace{-1em}}r|l@{ }l@{\hspace{-1em}}r@{}} \toprule
\multicolumn{2}{l}{\textbf{Labels}} & \textbf{\#spans}   & \multicolumn{2}{l}{\textbf{Labels}} & \textbf{\#spans} \\ \midrule
(1)  & \underline{Special deals} & 343 & (12) & \underline{Limited offers} & 52  \\
(2)  & \ \ Discount price    & 120      & (13) & \ \ Limited time       & 61  \\
(3)  & \ \ Reward points     & 85       & (14) & \ \ Limited target     & 114 \\
(4)  & \ \ Free              & 430      & (15) & \ \ First-time limited & 25  \\
(5)  & \ \ Special gift      & 126      & (16) & \underline{Track record} & 75  \\
(6)  & \underline{Features}  & 1,360    & (17) & \ \ Largest/no. 1      & 141 \\
(7)  & \ \ Quality           & 65       & (18) & \ \ Product lineup     & 258 \\
(8)  & \ \ Problem solving   & 17       & (19) & \ \ Trend              & 99  \\
(9)  & \ \ Speed             & 142      & (20) & \underline{Others}  & 182 \\
(10) & \ \ User-friendliness & 337      & (21) & \ \ Story              & 98  \\
(11) & \ \ Transportation    & 89       &      &                        &     \\ \bottomrule
\end{tabular}}
\caption{A$^3$ and statistics of annotated dataset, where ``\#spans'' represents the number of span texts annotated with each label.\label{tab:stats_of_annotation}}
\end{table}

We constructed a dataset of advertising appeals to understand the A$^3$ in ad texts.
Many A$^3$ exist in real-world advertisements, including product features, price, and campaigns.
We collected 782,158 ads from March 1, 2020 to February 28, 2021 through Google Ads,\footnote{\url{https://ads.google.com/}} which is an online advertising platform, to cover the expressions of advertising appeals in a wide range of industries.
In this work, we used ads in Japanese.
Each ad consists of a title, a description, and a landing page (LP),  which is a web page for a specific advertising campaign.
We used the meta-description\footnote{A meta-description is an HTML attribute that provides a brief summary of a web page, such as an LP.} of each LP as the LP content.
\ready{We sampled 5,000 ad texts 
for each advertiser to alleviate the bias owing to a different quantity of ad texts for the advertisers. 
Moreover, we excluded ad texts that comprised less than 15 characters or more than 200 characters.
The aforementioned two steps yielded 34,952 ad texts.
Furthermore, we excluded duplicates and highly similar texts using the normalized Levenshtein distance metric \cite{levenshtein1966bcc,greenhill-2011-levenshtein}, because the majority of the ad texts were created from templates for the sake of cost efficiency \cite{fujita2010-xm}}.
As a result, we collected 2,738 ad texts consisting of 666 titles, 1,532 descriptions, and 440 LP contents from 13 types of industries.\footnote{{\it EC}, {\it Media}, {\it Finance}, {\it VOD\&eBook}, {\it Cosmetics}, {\it Human resources}, {\it Education}, {\it Travel}, {\it Automobile}, {\it Entertainment}, {\it Real estate}, and {\it Beauty\&health}}
We provide the detailed statistics of the collected ad texts in Appendix \ref{sec:appendix_collected_dataset}.

\subsection{Label Types and Annotation Scheme}
Owing to the existence of various A$^3$, we believe that the systematic organization of the A$^3$ can aid the ad creation process.
We manually defined aspect labels in the following two phases.
First, we conducted a preliminary analysis of the collected ad texts and found that approximately eight aspects appeared: {\it special deals}, {\it quality}, {\it problem solving}, {\it speed}, {\it user-friendliness}, {\it limited offers}, {\it product lineup}, and {\it trend}.
Second, we presented these aspects and the collected ad texts to experienced ad creators and asked for their opinions on the A$^3$ with the aim of refining the aspect labels.
Consequently, the ad creators suggested that we further subdivide {\it special deals} and {\it limited offers}. 
For example,  {\it special deals} was subdivided into {\it discount price}, {\it reward points}, {\it free}, and {\it special gift}. 
The reason for this is that there are differences in the strength of the aspects between {\it free} and {\it special gift}, even though they appear to be similar.
Furthermore, {\it largest/no.1} was added as another aspect label because it attracts a lot of users.

Table~\ref{tab:stats_of_annotation} lists the A$^3$ that we manually defined.
Detailed descriptions and examples are provided in Appendix \ref{sec:appendix_descriptions_and_examples}.
Finally, we carefully designed a hierarchical scheme for A$^3$ to help ad creators and annotators to understand the differences between the labels.
The aspect hierarchy consists of five types of coarse-grained labels including {\it special deals}, which are underlined in Table \ref{tab:stats_of_annotation}, and 16 types of fine-grained labels such as {\it discount price}.

Because an ad text often contains multiple expressions of advertising appeals, as depicted in Figure \ref{fig:example_of_td}, we defined an advertising expression as a span text to be annotated.
For example, annotators provide the aspect label (e.g., {\it special deals}) for the span text ``{\it best price guarantee}.''
Each span was annotated during the annotation work. 
Moreover, we allowed the annotators to provide multiple labels for each span because an expression of advertising appeals may contain multiple aspects.
For example, the advertising expression ``{\it members get an extra 20\% off}'' contains two aspects {\it discount price} and {\it limited-target offer}, because it means that only users belonging to a membership program can receive an extra 20\% discount.

\subsection{Annotation Process}
We recruited six participants who worked at an advertising agency.
We separated 2,738 collected ad texts into two sets consisting of 1,100 and 1,638 texts, and assigned three participants to each set.
We presented a one-hour lecture to the participants to explain the detailed definitions of the labels and to provide annotation examples.
Furthermore, we asked them to annotate 30 ad texts that were separated from the collected dataset as a practice session.
After the session, we answered questions from the participants.
During the annotation, we answered any additional questions from them and shared information when a difficult case appeared, which was relatively rare.

\subsection{Annotated Dataset Statistics}

Table \ref{tab:stats_of_annotation} displays the statistics of the annotated dataset.
We adopted annotated spans only if at least two of the three annotators for each span text agreed with their boundaries and labels.
The annotation work for the 2,738 ad texts required a total of 42 hours; thus, the average time per ad text was 55.2 seconds.
A single ad text contains 1.54 spans on average.
Furthermore, we calculated the Cohen's Kappa coefficients ($\kappa$) between the tokens annotated by different pairs of annotators to determine the inter-annotator agreement. 
Moreover, following the previous work \cite{brandsen2020-ip}, we also report the $F_1$ scores that were calculated between the spans annotated by different pairs of annotators, where we considered one annotation as the ground truth and another as the prediction. 
We obtained relatively high agreement among the annotators: $\kappa=0.612, F_1=0.451$.


\section{Aspect Detection Model}
\ready{We investigate two existing models for aspect detection, i.e., the span-based \cite{zheng2019-a-boundary-aware} and document-based (doc-based) models \cite{devlin-etal-2019-bert}.}
These models receive an ad text $\bm{x}=(x_i)^{|\bm{x}|}_{i=1}$ as an input and predict aspect labels $\bm{y}={(y_i)}^{K}_{i=1}$, where $x_i$ and $y_i$ represent a token of an ad text and a binary label for each aspect label, respectively.
As each span may contain multiple aspects, both models perform label prediction in the form of multi-label classification~\cite{kurata-etal-2016-improved}.
$K$ is the number of aspect labels defined in Table \ref{tab:stats_of_annotation}.
We consider an expression of the advertising appeals in an ad text, such as ``{\it best price guarantee}'' in Figure \ref{fig:example_of_td}, to be a {\it span}.
We use $S(i,j)$ to represent the span from $i$ to $j$, where $1 \leq i < j \leq |\bm{x}|$.
The span-based model consists of two steps: (i) extracting a span $S(i,j)$ from $\bm{x}$ and (ii) predicting the aspect labels $\bm{y}$ for each span.
In contrast, the doc-based model predicts the aspect labels $\bm{y}$ for an entire ad text $\bm{x}$.
We employed a pre-trained BERT \cite{devlin-etal-2019-bert} for both models owing to the limited amount of the annotated dataset.

\subsection{Span-Based Model}
\begin{figure}[t]
\centering
\includegraphics[width=0.80\linewidth]{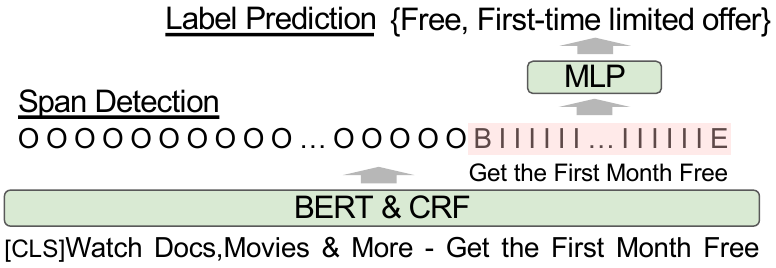}
\caption{Overview of the span-based model.\label{fig:span-based_aspect_detection}}
\end{figure}
Figure \ref{fig:span-based_aspect_detection} presents an overview of the span-based model.
\ready{The task of extracting a span from an ad text can be considered as named entity recognition, and we introduce the boundary-aware neural model proposed by \citet{zheng2019-a-boundary-aware}}.
We consider characters as a unit (token) in the span-based model.
We use the BIOE scheme to create boundary labels $\bm{l}=(l_i)^{|\bm{x}|}_{i=1}$ for the input tokens $\bm{x}$.
We feed $\bm{x}$ into the BERT to obtain a vector $h_i$ for $x_i$ for span detection.
Subseqently, we obtain the distribution of the boundary labels $v_i\in\mathbb{R}^L$ by applying a multilayer perceptron (MLP) $v_i=\mathtt{MLP}(h_i)$, where $L$ is the number of boundary types ($\mathtt{BIOE}$).
We also use a linear-chain conditional random field (CRF) \cite{lafferty2001-rz} to model the dependencies of the boundary labels (e.g., label ${\mathtt E}$ must appear after ${\mathtt B}$ or ${\mathtt I}$).
As a result, we can obtain the boundary labels $\bm{l}$ that are predicted by viterbi decoding for the input $\bm{x}$.

For label prediction, we create a vector representation $h^{avg}_{(i,j)}$ for a span $S(i,j)$ using the average of the output vectors of the BERT (i.e., $h_i, h_{i+1} \cdots h_j$). 
Thereafter, we obtain the probability that each span $S(i,j)$ belongs to the aspect labels $\bm{y}$ by applying an MLP and a sigmoid function $\bm{m}=\mathtt{Sigmoid}(\mathtt{MLP}(h^{avg}_{(i,j)}))$, where $\bm{m}=(m_k)^{K}_{k=1}$ and $m_k=p(y_k=1|S(i,j))$.
For example, in Figure \ref{fig:span-based_aspect_detection}, the expression ``{\it Get the First Month Free}'' is detected as a span, and the model predicts two aspect labels {\it free} and {\it first-time limited offer} for the detected span.

\subsection{Doc-Based Model\label{sec:doc-based_model}}
Although the span-based model offers the advantage of detecting a specific expression using span detection, we are concerned that errors in span detection could affect label prediction.
Therefore, we also introduce the doc-based model as an alternative to the span-based model.

The doc-based model is a BERT-based classification model. 
Following the original BERT-based classifier \cite{devlin-etal-2019-bert}, the doc-based model consists of a BERT and an MLP, which take an entire ad text $\bm{x}$ as an input and outputs labels $\bm{y}$.
Specifically, we first input the ad text $\bm{x}$ into the BERT and obtain the vector representation $h^{\mathtt{[CLS]}}$ for a {\tt [CLS]} token.
Subsequently, we feed the vector $h^{\mathtt{[CLS]}}$ into the MLP to obtain the probability that the ad text $\bm{x}$ belongs to the aspect labels $\bm{y}$ as a multi-label classification task $\bm{m}=\mathtt{Sigmoid}(\mathtt{MLP}(h^{\mathtt{[CLS]}}))$, where  $\bm{m}=(m_k)^{K}_{k=1}$ and $m_k=p(y_k=1|\bm{x})$.

\section{CTR Prediction with A$^3$}
\begin{figure}[t]
\centering
\includegraphics[width=0.80\linewidth]{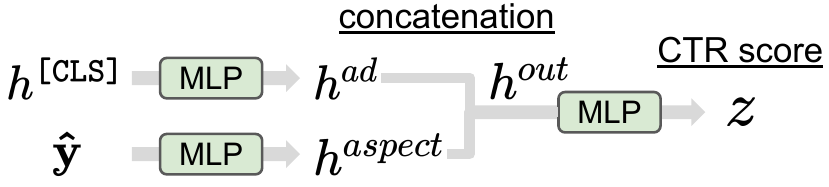}
\caption{Overview of the CTR prediction model.\label{fig:ctr_estimation_model}}
\end{figure}
Within the context of ad creation support, the estimation of advertising performance for an ad text (e.g., the CTR) plays a key role in both the improvement and cost efficiency of the ad creation because it helps us understand the user's interest.
Therefore, we also investigate whether the A$^3$ contributes to the prediction of the advertising performance.
For this task, we input an ad text $\bm{x}$ consisting of a title and description, an industry type of the ad $\bm{t}$ (e.g., EC), and keywords $\bm{k}$ (e.g., {\it tokyo} and {\it hotel}). 
We also introduce the predicted aspect labels $\bm{\hat{y}}$ (e.g., {\it features}) for $\bm{x}$ as additional features, which were detected by either the span-based or doc-based model.
In this case, we use the CTR $\bm{z}\in[0,1]$ as the advertising performance (CTR $=$ clicks $\div$ impressions).

Figure \ref{fig:ctr_estimation_model} presents an overview of the regression model.
Similarly to recent work \cite{mishra2021tsi}, we design this regression model based on the BERT.
In the model, we feed the three types of tokens $\bm{x}$, $\bm{t}$, $\bm{k}$ into the BERT to obtain the vector $h^{\mathtt{[CLS]}}$ for a {\tt [CLS]} token.
Subsequently, we input $h^{\mathtt{[CLS]}}$ and the aspect labels $\bm{\hat{y}}$ for the ad text $\bm{x}$ into the following MLP. 
Thereafter, we obtain the concatenated vector $h^{out}=[h^{ad};h^{aspect}]$, where ``$;$'' is a concatenation operator.
The final MLP then predicts a CTR score $\bm{z}$ from $h^{out}$ as $\bm{z}=\mathtt{Sigmoid}(\mathtt{MLP}(h^{out}))$.

\section{Experiments}
We conducted experiments on three tasks: (1) aspect detection for the A$^3$, (2) correlation analysis between the A$^3$ and CTR, and (3) CTR prediction.

\subsection{Experimental Settings}
\paragraph{Dataset}
We used the annotated dataset in Table \ref{tab:stats_of_annotation} for the aspect detection.
We separated the dataset into 1,857 samples for training, 465 for development, and 410 for testing after excluding 6 ad texts that we determined were inappropriately annotated. 
We collected 168,412 pairs of ad texts, keywords, and industry types from March 1, 2020 to February 28, 2021 through Google Ads for the CTR prediction.
We carefully separated the dataset into 136,352, 16,084, and 15,976 samples for training, development, and testing, respectively. 
The detailed statistics of the dataset for the CTR prediction are presented in Appendix \ref{sec:appendix_ctr_dataset}.
We used the training dataset for the CTR prediction for the correlation analysis between the CTR and A$^3$.
We used the campaign ID of each ad for data division to prevent leakage between the datasets.

\paragraph{Implementation}
We used the character-level BERT\footnote{\url{https://huggingface.co/cl-tohoku/bert-base-japanese-char}} for the span-based model, and the word-level BERT\footnote{\url{https://huggingface.co/cl-tohoku/bert-base-japanese}} for the doc-based model and CTR prediction.
We fine-tuned the models on the dataset and applied an early stopping strategy with 10 epochs.
The training was stopped if there was no improvement in the validation loss for three consecutive epochs in all experiments.
Further implementation details are described in Appendix \ref{sec:appendix_implementation}.

\paragraph{Evaluation Metrics}
We calculated the $F_1$ scores of the aspect labels for the aspect detection.
For the span-based model, a detected label was considered as a true positive if both its span and label were correctly detected.
We used the area under the receiver operating characteristic curve (AUC) \cite{fawcett2006-ab}, which is a widely used metric in the field of CTR prediction \cite{zhou2018-oj,xiao2020-kx}.
Moreover, we used the root-mean-squared error (RMSE) and mean absolute error (MAE) to measure the differences between the ground-truth and predicted scores.

\subsection{Aspect Detection}
\renewcommand{\arraystretch}{0.95}
\begin{table}[t]
\centering
{\small
\begin{tabular}{@{}c@{ }ll@{}r@{\hspace{1.0em}}r@{\hspace{1.2em}}r@{\hspace{1em}}@{}} \toprule
\multicolumn{3}{c}{\multirow{2}{*}{\textbf{Labels}}} & \multicolumn{2}{c}{\textbf{Span-based}} & \multicolumn{1}{c}{\textbf{Doc-}} \\ 
     & \multicolumn{2}{l}{} & \multicolumn{1}{l}{\texttt{Pred}} & \texttt{Orac} & \multicolumn{1}{c}{\textbf{based}}     \\ \midrule
(1)  & \multicolumn{2}{l}{Special deals} & 0.11 & 0.19 & 0.70 \\
(2)  &             & Discount price       & 0.00 & 0.00 & 0.57 \\
(3)  &             & Reward points        & 0.62 & 0.74 & 0.75 \\
(4)  &             & Free                 & 0.68 & 0.88 & 0.94 \\
(5)  &             & Special gift         & 0.28 & 0.40 & 0.65 \\ \midrule
(6)  & \multicolumn{2}{l}{Features}       & 0.50 & 0.70 & 0.72 \\
(7)  &             & Quality              & 0.00 & 0.00 & 0.44 \\
(8)  &             & Problem solving      & 0.00 & 0.00 & 0.00 \\
(9)  &             & Speed                & 0.51 & 0.66 & 0.92 \\
(10) &             & User-friendliness    & 0.46 & 0.59 & 0.56 \\
(11) &             & Transportation       & 0.91 & 1.00 & 0.53 \\ \midrule
(12) & \multicolumn{2}{l}{Limited offers} & 0.38 & 0.53 & 0.62 \\
(13) &             & Limited time         & 0.00 & 0.00 & 0.47 \\
(14) &             & Limited target       & 0.26 & 0.57 & 0.44 \\
(15) &             & First-time limited   & 0.00 & 0.00 & 0.00 \\ \midrule
(16) & \multicolumn{2}{l}{Performance}    & 0.27 & 0.50 & 0.48 \\
(17) &             & Largest/no. 1        & 0.67 & 0.80 & 0.82 \\
(18) &             & Product lineup       & 0.42 & 0.67 & 0.67 \\
(19) &             & Trend                & 0.41 & 0.56 & 0.47 \\ \midrule
(20) & \multicolumn{2}{l}{Others}         & 0.00 & 0.00 & 0.39 \\
(21) &             & Story                & 0.32 & 0.83 & 0.53 \\ \midrule
\multicolumn{3}{c}{\textbf{Macro average}}& 0.32 & 0.46 & 0.56 \\ \bottomrule
\end{tabular}}
\caption{Results of the aspect detection ($F_1$ scores)\label{tab:result_of_aspect_detection}}
\end{table}
\renewcommand{\arraystretch}{1.0}

In this experiment, we evaluated two models, the span-based and doc-based models. 
As errors in the span prediction may affect the label prediction in the span-based model, we also introduced the {\tt Oracle} model, which predicts their labels, provided with {\it oracle} spans, in addition to the {\tt Pred} model, which predicts both the spans and labels.

The evaluation results for the aspect detection are presented in Table \ref{tab:result_of_aspect_detection}.
The doc-based model outperformed the span-based model, including the 
{\tt Oracle} model, for most A$^3$.
As the {\tt Pred} model is required to predict both the spans and labels correctly, its task is relatively more difficult than that of other models.
In fact, we found that the $F_1$ score for the span detection is 0.69 for the {\tt Pred} model.
Therefore, we conclude that it is the reason why the macro-average $F_1$ score of {\tt Pred} was lower than those of the doc-based and {\tt Oracle} models. 

\ready{In the comparison between the {\tt Oracle} and doc-based models, the doc-based model outperforms the {\tt Oracle} model.
We hypothesize that its training objective for the span-based model is more difficult as it is more fine grained than the doc-based model.}

We observed that the scores for {\it free}, {\it speed}, and {\it largest/no. 1} are high 
in the doc-based model.
This implies that the advertising expressions for these aspects are relatively monotonous and easy to detect compared to the other aspects.
For example, the advertising expression ``{\it free shipping},'' which belongs to {\it free}, often occurs frequently in ad texts for a wide range of industries.
The aspect detection was difficult for several aspects in which the numbers of annotated cases were limited, such as (8) and (15), as indicated from Tables \ref{tab:stats_of_annotation} and \ref{tab:result_of_aspect_detection}.
\ready{Hence, they exhibited an $F_1$ score of 0.00.}

\begin{figure}[t]
\centering
  \begin{minipage}[b]{0.49\linewidth}
    \centering
    \includegraphics[width=1.00\linewidth]{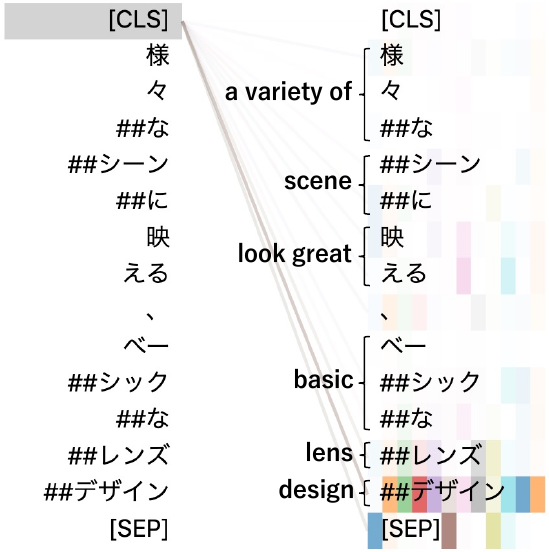}
    {\scriptsize {\it ``Simple-designed eyeglass lenses that look great in a variety of scenes''}}
    \subcaption{{\scriptsize Ad text labeled as {\it features}}}
  \end{minipage}
  \begin{minipage}[b]{0.49\linewidth}
    \centering
    \includegraphics[width=1.00\linewidth]{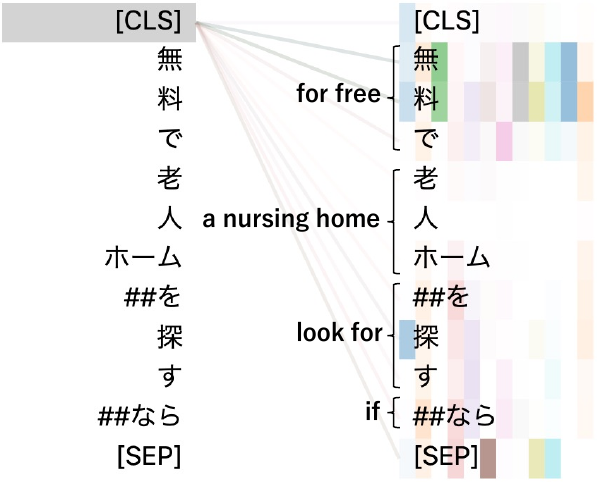}
    {\scriptsize {\it ``Find a nursing home for free''}}
    \subcaption{{\scriptsize Ad text labeled as {\it free}}}
  \end{minipage}
\caption{\draft{Visualization of attention weights in the doc-based model. Each example consists of the original Japanese ad text with the literal translation for each subword and the corresponding English ad text. \label{fig:attention_analysis}}}
\end{figure}
We also conducted an analysis of the attention in the doc-based model to understand to which signals the model attended in the aspect detection.
Figure \ref{fig:attention_analysis} depicts the visualized attention patterns with respect to the {\tt [CLS]} token of the final layer of the BERT. 
We found that many of the attention heads attend to the words ``{\it design}'' and ``{\it for free}'' for the ad text (a) and (b), respectively.
This suggests that the doc-based model classified the ad text (a) and (b) as {\it features} and {\it free}, respectively, because these words were related to the aspects.

\subsection{Correlation between Aspects and CTR}
\begin{table}[t]
\centering
\renewcommand{\arraystretch}{0.92}
\begin{small}
\begin{tabular}{@{}c@{ }r@{ }r@{ }r@{ }r@{ }r@{}} \toprule
\multicolumn{1}{c}{\textbf{Labels}}  & \multicolumn{1}{c}{\textbf{eBook}} & \multicolumn{1}{c}{\textbf{EC}} & \multicolumn{1}{c}{\textbf{Fin}} & \multicolumn{1}{c}{\textbf{HR}} & \multicolumn{1}{c}{\textbf{Travel}}     \\ \midrule
(1)  & 0.229  & 0.011  & -0.171 & \multicolumn{1}{c}{$-$} & 0.017\\
(2)  & -0.135 & -0.166 & -0.128 & \multicolumn{1}{c}{$-$} & -0.176\\
(3)  & 0.183  & 0.000  & {\bf 0.443} & \multicolumn{1}{c}{$-$} & {\bf 0.377}\\
(4)  & -0.126 & -0.163 & -0.052 & 0.116  & \multicolumn{1}{c}{$-$}\\
(5)  & 0.086  & 0.122  & {\bf 0.339} & -0.024 & {\bf -0.332}\\ \midrule
(6)  & -0.128 & -0.121 & -0.094 & -0.040 & 0.050\\
(7)  & -0.001 & -0.081 & -0.034 & \multicolumn{1}{c}{$-$}  & \multicolumn{1}{c}{$-$}\\
(8)  & \multicolumn{1}{c}{$-$} & \multicolumn{1}{c}{$-$}  & \multicolumn{1}{c}{$-$} & \multicolumn{1}{c}{$-$}  & \multicolumn{1}{c}{$-$}\\
(9)  & -0.017 & 0.065 & -0.109 & 0.024 & \multicolumn{1}{c}{$-$}\\
(10) & -0.236 & 0.053 & {\bf -0.252} & -0.004 & 0.205\\
(11) & \multicolumn{1}{c}{$-$} & \multicolumn{1}{c}{$-$}  & \multicolumn{1}{c}{$-$} & \multicolumn{1}{c}{$-$}  & \multicolumn{1}{c}{$-$}\\ \midrule
(12) & -0.036 & -0.149 & -0.044 & 0.003   & 0.221\\
(13) & -0.090 & 0.186  & 0.014  & -0.006  & -0.184\\
(14) & -0.020 & -0.162 & -0.011 & 0.023   & \multicolumn{1}{c}{$-$}\\
(15) & -0.165 & \multicolumn{1}{c}{$-$}  & \multicolumn{1}{c}{$-$}   & \multicolumn{1}{c}{$-$}  & \multicolumn{1}{c}{$-$}\\ \midrule
(16) & 0.108  & -0.161 & -0.099   & 0.237  & -0.148\\
(17) & {\bf 0.283} & -0.073   & 0.143    & 0.102  & \multicolumn{1}{c}{$-$}\\
(18) & -0.206 & 0.044  & -0.005 & -0.159 & -0.195\\ 
(19) & -0.074 & -0.007 & 0.157  & \multicolumn{1}{c}{$-$} & \multicolumn{1}{c}{$-$}\\ \midrule
(20) & 0.022  & -0.083 & 0.134  & -0.042 & {\bf 0.268}\\
(21) & -0.093 & \multicolumn{1}{c}{$-$}  & \multicolumn{1}{c}{$-$} & \multicolumn{1}{c}{$-$}  & \multicolumn{1}{c}{$-$}\\ \midrule
\multicolumn{1}{c}{\textbf{\#cases}} & 30,536 & 20,671 & 20,183 & 10,823 & 8,093 \\ \bottomrule
\end{tabular}
\end{small}
\caption{Point-biserial correlation coefficient ${\bf r}$, where ``\# cases'' denotes the number of ad texts for each industry type and ``$-$'' indicates that the corresponding labels were not found. \label{tab:correlation_between_aspect_and_ctr}}
\renewcommand{\arraystretch}{1}
\end{table}
To realize the ad creation process considering the A$^3$, we analyzed which A$^3$ were effective in each industry through correlation analysis between the CTR\footnote{We used the {\it actual} CTR for each ad rather than the {\it predicted} CTR.} and the aspect labels that were predicted by the doc-based model.
Because the aspect labels are binary for each aspect (e.g., whether or not each aspect is included in an ad text) and the CTR is continuous, we used the point-biserial correlation coefficient ${\bf r}$ for the analysis. 
Table \ref{tab:correlation_between_aspect_and_ctr} lists the point-biserial correlation coefficients ${\bf r}$ between the aspect labels and the CTR.
We investigated the correlation among the industry types {\it VOD\&eBook (eBook)}, {\it EC}, {\it Finance (Fin)}, {\it Human resources (HR)}, and {\it Travel}. 
As indicated in \textbf{bold text} in Table \ref{tab:correlation_between_aspect_and_ctr}, we observed a weak correlation (0.25~<~$|{\bf r}|$~<~0.5) between the CTR and the labels, such as (3) {\it reward points} for {\it Finance}.
This implies that ad texts that include effective A$^3$ tend to attract more attention from users.
\ready{However, there was no correlation with regard to the other aspects.
This may be because (1) {\it features}, for example, is considered to be a general-purpose aspect and can be used in any situation.}

Based on the above insights, we also investigated the expressions for the effective A$^3$ in our annotated dataset.
For example, regarding the {\it VOD\&eBook} industry, we found that the expression ``{\it one of the largest websites in Japan}'' (\ja{国内最大級サイト}) was annotated as (17) {\it largest/no. 1}. 
Furthermore, the expressions for {\it Finance} ``{\it get {\tt [N]} points for new membership}'' (\ja{新規入会＆利用で{\tt [N]}ポイント}) and ``{\it earn {\tt [N]} points per {\tt [N]} yen}'' (\ja{{\tt [N]}円につき{\tt [N]}ポイント貯まる}) were labeled with (3) {\it reward points}.\footnote{Numbers (e.g., price, points) are masked with {\tt [N]}.}
We believe that the presentation of these effective expressions to ad creators may provide actionable insights and aid in the ad creation process. 

\subsection{CTR Prediction}

\begin{table}[t]
\centering
\begin{small}
\begin{tabular}{lrrr} \toprule
     & \multicolumn{1}{c}{\textbf{AUC} ($\uparrow$)} & \multicolumn{1}{c}{\textbf{RMSE} ($\downarrow$)} & \multicolumn{1}{c}{\textbf{MAE} ($\downarrow$)} \\ \midrule
BERT                        & 0.683          & 0.220  & 0.142  \\
\ \ \ + $\mathbf{l}_{span}$ & 0.709          & 0.218 &  0.137 \\
\ \ \ + $\mathbf{l}_{doc}$  & \textbf{0.713} & \textbf{0.217}  & \textbf{0.136} \\ \bottomrule
\end{tabular}
\end{small}
\caption{Results of CTR prediction\label{tab:result_of_estimation}}
\end{table}
We investigated whether the identification of the A$^3$ contributes to the estimation accuracy of the CTR.
Table \ref{tab:result_of_estimation} presents the results of the CTR prediction.
For comparison with a baseline (BERT), that does not use A$^3$, we introduced two models that consider A$^3$ predicted by the span-based model (+$\mathbf{l}_{span}$) or the doc-based model (+$\mathbf{l}_{doc}$).
It can be observed that the aspect-aware models that leverage the A$^3$ outperformed the baseline model in terms of all evaluation metrics.
This suggests that the identification of the A$^3$ that are included in ad texts can contribute to the improvement of CTR prediction.
In the comparison between the two models, +$\mathbf{l}_{doc}$ improved the performance of the CTR prediction more than the +$\mathbf{l}_{span}$.
This is likely because the doc-based model predicted the aspect labels more accurately than the span-based model, as indicated in Table \ref{tab:result_of_aspect_detection}.
We believe that improving the aspect detection with more refined methods will lead to better correlation and prediction for the CTR.

\section{Conclusions}
In this work, we have explored the effective A$^3$ by means of aspect detection and correlation analysis towards ad creation support with the A$^3$. 
Our experimental results demonstrated that each industry exhibits unique effective A$^3$ and that identification of the A$^3$ can contributes to CTR prediction.

\ready{We demonstrate two possible directions for future studies.
First, we will investigate whether introducing the effective A$^3$ in the ad creation process can help ad creators write effective ad texts in real-world applications.
Second, we will develop an aspect-aware model to automatically generate ad texts to support the ad creation process. 
For the latter, we will train the model with a dataset that includes pairs of ad texts and their corresponding aspect labels predicted using aspect detection.}

\bibliography{custom}
\bibliographystyle{acl_natbib}

\appendix
\clearpage

\begin{table*}[th]
\centering
{\small
\begin{tabular}{@{}c@{ }lll@{ }r@{}} \toprule
\multicolumn{3}{c}{\textbf{Aspect labels}}& \multicolumn{1}{c}{\textbf{Description \& Example}} & \textbf{\#spans} \\ \midrule
(1)  & \multicolumn{2}{l}{Special deals}  & Expressions representing special deals (e.g., {\it Compare hotels and save money}) & 343   \\
(2)  &      & Discount price              & Specific discount rate or amount (e.g., {\it Buy 1 get 1 50\% off})                & 120   \\
(3)  &      & Reward points               & Customers can earn points (e.g., {\it Use our app to earn points})                 & 85   \\
(4)  &      & Free                        & Free offer for products or services (e.g., {\it Enjoy free shipping})              & 430   \\
(5)  &      & Special gift                & Special gifts or presents for customers (e.g., {\it Join today and get a free brush set}) & 126   \\ \midrule
(6)  & \multicolumn{2}{l}{Features}       & Features of services or products  (e.g., {\it Ergonomically designed to protect children})       & 1,360   \\
(7)  &      & Quality                     & Top-quality or high-grade services (e.g., {\it Find premium kitchen appliances}) & 65   \\
(8)  &      & Problem solving             & Solutions to customer problems (e.g., {\it Get bright, clear skin})           & 17   \\
(9)  &      & Speed                       & Speed of delivery and services (e.g., {\it Fast \& free shipping})           & 142   \\
(10) &      & User friendliness           & Usability of services and products  (e.g., {\it Quick, simple, and easy to use })     & 337   \\
(11) &      & Transportation              & Convenience of transportation  (e.g., {\it Centrally located in the heart of Tokyo}) & 89   \\ \midrule
(12) & \multicolumn{2}{l}{Limited offers} & Limited availability of services and products  (e.g., {\it Limited to 1,000 items per day})  & 52   \\
(13) &      & Limited-time offer          & Offers available for a limited time only (e.g., {\it Three days only at 20\% off}) & 61   \\
(14) &      & Limited-target offer        & Offers available for target customers only  (e.g., {\it Discount for members only}) & 114   \\
(15) &      & First-time limited offer    & Limited offers for first-time customers (e.g., {\it Take 15\% off your first order})  & 25   \\ \midrule
(16) & \multicolumn{2}{l}{Track record}   & Track records of services or companies  (e.g., {\it 45M+ users worldwide}) & 75   \\
(17) &      & Largest/no. 1               & Largest/No. 1 products or services  (e.g., {\it Boston's no. 1 hair salon}) & 141   \\
(18) &      & Product lineup              & Wide range of products or stores (e.g., {\it Large selection of hotels})   & 258   \\
(19) &      & Trend                       & Popularity or favorable reputation  (e.g., {\it Top trending shoes and boots}) & 99   \\ \midrule
(20) & \multicolumn{2}{l}{Others}         & Other advertising appeals  (e.g., {\it An experience like no other}) & 182   \\
(21) &      & Story                       & Synopsis of a movie or drama (e.g., {\it After Peter Parker is bitten by a$\cdots$} )         & 98   \\ \bottomrule
\end{tabular}}
\caption{A$^3$ and statistics of annotated dataset. \label{tab:label_definition}}
\end{table*}

\begin{table*}[th]
\centering
{\small
\begin{tabular}{@{}l@{ }r@{\hspace{0.7em}}r@{\hspace{0.7em}}r@{}}
\toprule
 & \multicolumn{2}{c}{\textbf{Aspect Detection Model}} & \multicolumn{1}{c}{\multirow{2}{*}{\textbf{CTR Prediction Model}}} \\ \cmidrule(r){2-3}
 & \multicolumn{1}{c}{\textbf{Span-based}} & \multicolumn{1}{c}{\textbf{Doc-based}} &  \\ \midrule
Pre-trained model & {\footnotesize {\tt bert-base-japanese-char}} & {\footnotesize {\tt bert-base-japanese}} & {\footnotesize {\tt bert-base-japanese}} \\
Number of heads & 12 & 12 & 12 \\
Number of hidden layers & 12 & 12 & 12 \\
Hidden layer size & 768 & 768 & 768 \\
Dropout probability & 0.1 & 0.1 & 0.1 \\
Vocab size & 4,000 & 32,000 & 32,000 \\
Batch size & 10 & 10 & 30 \\
Max sequence length & 512 & 512 & 512 \\
Number of epochs & 10 & 10 & 10 \\
Learning rate & $8.6\times10^{-5}$ & $5.5\times10^{-5}$ & $2.0\times10^{-5}$ \\
Optimizer & Adam & Adam & Adamax \\
Loss & CRF loss, BCE loss & BCE loss & MSE loss\\ \bottomrule
\end{tabular}}
\caption{Hyperparameters and implementation details. \label{tab:hyperparams}}
\end{table*}

\begin{table}[t]
\renewcommand{\arraystretch}{0.98}
\centering
\begin{small}
\begin{tabular}{@{}l@{\hspace{0.7em}}r@{\hspace{1.0em}}r@{\hspace{1.0em}}r@{\hspace{1.0em}}r@{}} \toprule
\textbf{Industry} & \textbf{Title} & \textbf{Desc.} & \textbf{LP}   & \textbf{Sub-total} \\ \midrule
EC               & 131   & 314   & 87   & 532   \\ 
Others           & 137   & 272   & 123  & 532   \\  
Media            & 119   & 250   & 27   & 396   \\ 
Finance          & 105   & 203   & 56   & 364   \\ 
VOD\&eBook       & 38    & 112   & 78   & 228   \\ 
Cosmetics        & 43    & 110   & 20   & 173   \\ 
Human resources  & 72    & 75    & 8    & 155   \\ 
Education        & 58    & 50    & 10   & 118   \\ 
Travel           & 23    & 62    & 18   & 103   \\ 
Automobile       & 18    & 32    & 5    & 55    \\ 
Entertainment    & 14    & 36    & 3    & 53    \\ 
Real estate      & 5     & 12    & 2    & 19    \\ 
Beauty\&Health   & 3     & 4     & 3    & 10    \\ \midrule 
Total            & 766   & 1,532 & 440  & 2,738 \\ \bottomrule 
\end{tabular}
\end{small}
\caption{\draft{Statistics of collected ad texts.}\label{tab:stats_of_dataset}}
\renewcommand{\arraystretch}{1.00}
\end{table}

\begin{table}[t]
\centering
\begin{small}
\begin{tabular}{lrrr} \toprule
\multicolumn{1}{c}{Industry} & \multicolumn{1}{c}{\textbf{Train}} & \multicolumn{1}{c}{\textbf{Dev}} & \multicolumn{1}{c}{\textbf{Test}} \\ \midrule
VOD\&eBook        & 30,536  & 3,823  & 3,812  \\
EC                & 20,671  & 2,584  & 2,583  \\
Finance           & 20,183  & 2,521  & 2,521  \\
Others            & 15,526  & 1,936  & 1,936  \\
Human resources    & 10,823  & 1,348  & 1,348  \\
Media             & 10,434  & 1,295  & 1,274  \\
Education         & 9,592   & 1,344  & 1,228  \\
Travel            & 8,093   & 1,002  & 1,042  \\
Cosmetics         & 5,584   & 231    & 232    \\
Entertainment     & 2,455   & 0      & 0      \\
Automobile        & 1,697   & 0      & 0      \\
Beauty\&Health    & 445     & 0      & 0      \\
Real estate       & 313     & 0      & 0      \\ \midrule
Total             & 136,352 & 16,084 & 15,976 \\ \bottomrule
\end{tabular}
\end{small}
\caption{\draft{Statistics of dataset for CTR prediction.\label{tab:dataset_for_ctr_prediction}}}
\end{table}
\ready{
\section{Collected Ad Texts for Annotation \label{sec:appendix_collected_dataset}}
Table \ref{tab:stats_of_dataset} lists the detailed statistics of the collected ad text.
We collected 2,738 ad texts comprising 666 titles $\bm{x}^{title}$, 1,532 descriptions $\bm{x}^{desc}$, and 440 LP contents $\bm{x}^{lp}$ from 13 industries: {\it EC}, {\it Media}, {\it Finance}, {\it VOD\&eBook}, {\it Cosmetics}, {\it Human resources}, {\it Education}, {\it Travel}, {\it Automobile}, {\it Entertainment}, {\it Real estate}, and {\it Beauty\&Health}.}

\ready{
\section{Descriptions and Examples of A$^3$ \label{sec:appendix_descriptions_and_examples}}
Table \ref{tab:label_definition} lists the detailed descriptions and examples of A$^3$ that we have defined.
For example, the expression ``{\it enjoy free shipping}'' is labeled with (4) {\it free}, as it represents free offers for products or services.
In the table, ``\#spans'' represents the number of span texts annotated with each label.}

\ready{
\section{Dataset for CTR Prediction \label{sec:appendix_ctr_dataset}}
Table \ref{tab:dataset_for_ctr_prediction} lists the detailed statistics of the datasets used for CTR prediction.
We carefully separated the dataset into 136,352, 16,084, and 15,976 samples for training, development, and testing, respectively. 
For correlation analysis between the CTR and aspect labels of advertising appeals, we used the training dataset for CTR prediction.}

\ready{
\section{Additional Implementation Details \label{sec:appendix_implementation}}
\ready{Table \ref{tab:hyperparams}} lists the implementation details, e.g., hyperparameters, for the aspect detection and CTR prediction models.
We developed our models using pre-trained BERT models, which are publicly available from the Transformers library \cite{wolf-etal-2020-transformers}.\footnote{\url{https://huggingface.co/cl-tohoku}}
The framework is available under the Apache 2.0 license.
We trained the models with a Tesla V100 GPU on the Google Cloud Platform, which is the cloud computing infrastructure.
Moreover, we performed a hyperparameter search, using Optuna \cite{optuna_2019} with default parameters for the aspect detection models on the validation set. 
In the experiment, the hyperparameter search is limited to 30 trials.
Therefore, we performed our experiments in a single run.}

\ready{
We used CRF and binary cross-entropy (BCE) loss for span detection and label prediction in the span-based model, respectively. 
We used the mean squared error (MSE) as an objective function to train the CTR prediction model.
Furthermore, we applied an early stopping strategy to all the models. 
Specifically, we stopped training if there was no improvement in the validation loss after three consecutive epochs.}

\end{document}